\documentclass[runningheads]{llncs}
\usepackage[T1]{fontenc}
\usepackage{graphicx}
\usepackage{cite}
\usepackage{amsmath,amssymb,amsfonts}
\usepackage{algorithmic}
\usepackage{textcomp}
\usepackage{xcolor}

\usepackage{times}
\usepackage{soul}
\usepackage{url}
\usepackage[hidelinks]{hyperref}
\usepackage[utf8]{inputenc}
\usepackage{caption}
\usepackage{booktabs}
\usepackage{algorithm}
\usepackage[switch]{lineno}
\usepackage{amsfonts}
\usepackage{subfig}
\usepackage{enumitem}
\usepackage{multirow}

\usepackage{tikz}

\soulregister\cite7 
\soulregister\citep7 
\soulregister\citet7 
\soulregister\ref7 
\soulregister\pageref7 
\begin{document}
\title{Beyond Random Missingness: Clinically Rethinking for Healthcare Time Series Imputation}
\author{Linglong Qian\inst{1,2} \and
Yiyuan Yang\inst{2,3} \and
Wenjie Du\inst{2} \and
Jun Wang\inst{2,4} \and
Richard Dobsoni\inst{1,5} \and
Zina Ibrahim\inst{1,2} \thanks{Corresponding author.}
}
\authorrunning{L. Qian et al.}
\institute{King's College London, London, United Kingdom \and
PyPOTS Research, China \and
University of Oxford, Oxfordshire, United Kingdom \and
Hong Kong University of Science and Technology, Hong Kong, China \and
University College London, London, United Kingdom
\email{\{linglong.qian,richard.j.dobson,zina.ibrahim\}@kcl.ac.uk, yiyuan.yang@cs.ox.ac.uk, wdu@pypots.com, jwangfx@connect.ust.hk}}

\maketitle              % typeset the header of the contribution
\begin{abstract}
This study investigates the impact of masking strategies on time series imputation models in healthcare settings. While current approaches predominantly rely on random masking for model evaluation, this practice fails to capture the structured nature of missing patterns in clinical data. Using the PhysioNet Challenge 2012 dataset, we analyse how different masking implementations affect both imputation accuracy and downstream clinical predictions across eleven imputation methods. Our results demonstrate that masking choices significantly influence model performance, while recurrent architectures show more consistent performance across strategies. Analysis of downstream mortality prediction reveals that imputation accuracy doesn't necessarily translate to optimal clinical prediction capabilities. Our findings emphasise the need for clinically-informed masking strategies that better reflect real-world missing patterns in healthcare data, suggesting current evaluation frameworks may need reconsideration for reliable clinical deployment. The source code is available at \url{https://github.com/LinglongQian/Mask_rethinking}.
\end{abstract}

\section{Introduction}
The increasing digitisation of healthcare through Electronic Health Records (EHRs) has created unprecedented opportunities for developing predictive algorithms to support clinical decision-making \cite{wahl2018artificial}. However, EHR data presents unique challenges for machine learning applications, particularly in handling missing values, which are pervasive and often carry clinical significance \cite{wells2013strategies}. Unlike many domains where data completeness can be controlled, missingness in healthcare data is intrinsically linked to clinical workflows, patient conditions, and institutional protocols \cite{garcia2010pattern}.

Recent comprehensive studies, particularly the TSI-Bench framework \cite{tsi2024} and systematic evaluations of deep imputation models \cite{qian2024deep}, have revealed critical gaps between theoretical model capabilities and practical performance in handling healthcare time series. These studies highlight that while modern deep learning approaches demonstrate promising results in controlled settings, their evaluation frameworks often fail to capture the complex nature of missingness in clinical data. This misalignment is particularly concerning as imputation quality directly impacts downstream clinical prediction tasks \cite{che2018recurrent}.

In clinical settings, missing data patterns exhibit sophisticated dependencies that reflect both medical decision-making and patient care trajectories. For instance, vital signs monitoring frequency typically increases during critical care episodes, creating temporal clusters of observations, while routine laboratory tests follow institution-specific protocols, generating structured missing patterns. Traditional evaluation approaches using random masking fail to capture these nuanced relationships, potentially leading to overoptimistic performance estimates when models are deployed in real clinical environments.

The implications of masking strategies extend beyond model evaluation to impact clinical reliability. Consider a scenario where a patient's deterioration is indicated by subtle changes in multiple physiological parameters. The choice between temporal masking (simulating equipment disconnection) versus block masking (mimicking comprehensive data loss during patient transfer) can significantly affect a model's ability to capture such critical patterns. Similarly, the timing of data normalisation relative to masking can influence how well a model learns to handle the extreme values often indicative of acute clinical events.

This study presents a systematic investigation of how different masking strategies affect the performance of deep imputation models in healthcare settings. Our findings demonstrate that the common practice of random masking may inadequately prepare models for real-world clinical applications. We advocate for more sophisticated, clinically informed masking strategies that better reflect the structured nature of missing patterns in healthcare data. This work contributes to the broader discussion of evaluation methodologies in healthcare AI, emphasising the need for evaluation frameworks that align with the practical challenges of clinical data analysis.

\begin{figure*}[tbp]
\centering
\resizebox{1.0\textwidth}{!}{
\subfloat[Random Masking]{\includegraphics[width=3cm]{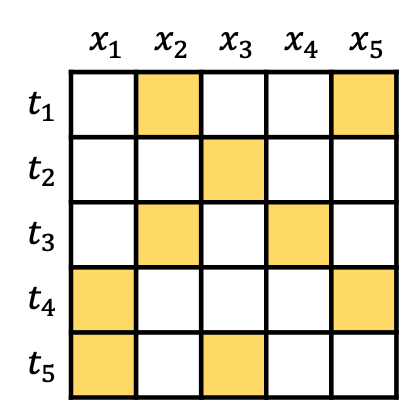}}
\subfloat[Temporal Masking]{\includegraphics[width=3cm]{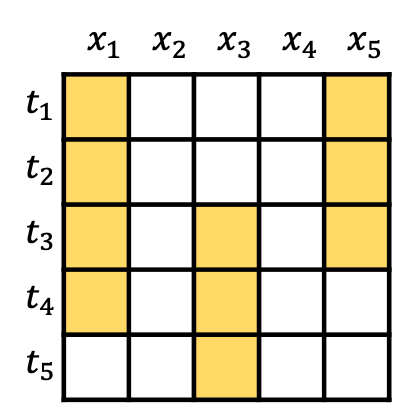}}
\subfloat[Spatial Masking]{\includegraphics[width=3cm]{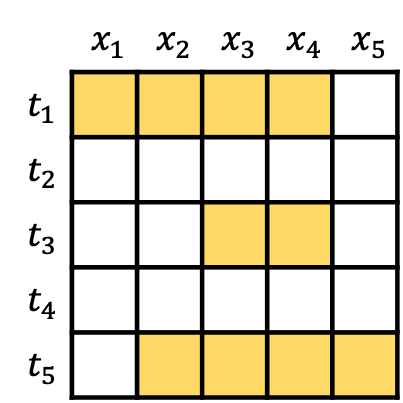}} 
\subfloat[Block Masking]{\includegraphics[width=3cm]{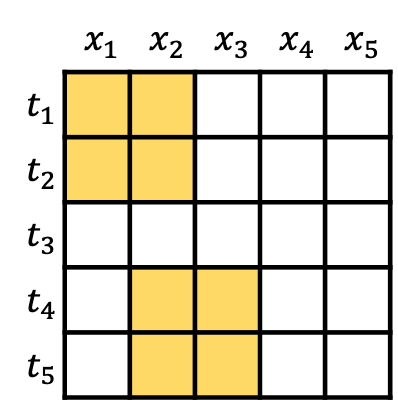}}
\subfloat[Augmentation]{\includegraphics[width=3cm]{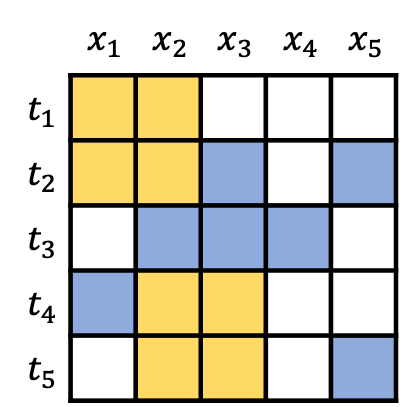}}
\subfloat[Overlaying]{\includegraphics[width=3cm]{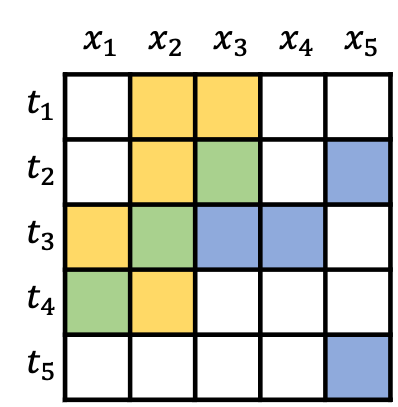}}
}
\caption{Masking techniques and approaches demonstrated over a time-series of five features ($x_1 \sim x_5$) and five-time points ($t_1 \sim t_5$): (a) random masking, (b) temporal masking, (c) spatial masking, (d) block masking. The yellow cells indicate those labelled as missing via masking. In (e) augmentation and (f) overlaying, the blue cells indicate cells that are missing within the original data. In (e), the masked (yellow) cells have no overlap with the original missingness in the data. Green: masked data coming from both the original missingness and artificial missingness. In (f), overlaying masks cells from either the original missingness or simulates artificial missingness from non-missing data.} \label{fig:masking}
\end{figure*}

\section{Related Work}
The landscape of missing data imputation in healthcare has evolved significantly with the advent of deep learning approaches. Our review organises the literature along two key dimensions: the architectural approaches to handling medical time series and their alignment with healthcare-specific missing patterns.

\subsection{Deep Learning Architectures for Medical Time Series}
Recent comprehensive reviews \cite{reviewhealthcare2,wang2024deep} highlight the diversity of neural architectures employed for medical time series imputation, each making distinct assumptions about the nature of healthcare data. 
% As shown in Table \ref{tab:overview1}, these approaches can be broadly categorised by their fundamental architectural choices and the types of missingness they are designed to handle.

\paragraph{RNN-based Approaches} Early adaptations of recurrent architectures for healthcare data demonstrated the importance of capturing temporal dependencies in clinical measurements. GRUD \cite{che2018recurrent} pioneered the incorporation of medical domain knowledge by introducing exponential decay to model the diminishing relevance of past clinical observations, particularly relevant for vital sign monitoring where both natural physiological decay and equipment-based missingness (e.g., patient disconnection during procedures) occur. BRITS \cite{cao2018brits} advanced this approach by incorporating bidirectional dynamics and explicit modelling of feature relationships, addressing the complex interdependencies between clinical variables such as the correlation between blood pressure measurements and heart rate.

\paragraph{Attention-based Models} The emergence of attention mechanisms has enabled more sophisticated modelling of long-term clinical dependencies. SAITS \cite{du2023saits} employs dual-view attention to capture both temporal patterns within individual vital signs and cross-variable interactions, particularly valuable for detecting subtle clinical deterioration patterns that manifest across multiple parameters. Similarly, MTSIT \cite{yildiz2022multivariate} introduces position-aware attention specifically designed to handle the irregular sampling intervals common in clinical data collection.

\paragraph{Probabilistic Approaches} Models incorporating explicit uncertainty quantification have gained prominence due to the critical nature of medical decisions. GP-VAE \cite{fortuin2020gp} combines Gaussian processes with variational autoencoders to model the uncertainty in clinical measurements, while CSDI \cite{tashiro2021csdi} employs diffusion models to generate realistic physiological trajectories. These approaches are particularly relevant for critical care settings, where confidence in imputed values directly impacts clinical decision-making.

\subsection{Alignment with Clinical Missing Patterns}
The evaluation of these models reveals a significant gap between their theoretical capabilities and practical performance in healthcare settings. As detailed in \cite{qian2024deep}, most models claim to handle various missingness mechanisms (MCAR, MAR, MNAR) but are predominantly evaluated using random masking strategies that fail to capture the structured nature of clinical data collection.

\paragraph{Clinical Missing Mechanisms} Healthcare data exhibits distinct missing patterns reflective of clinical workflows:
\begin{itemize}
    \item \textbf{Protocol-Driven Missingness:} Regular vital sign measurements following nursing schedules (e.g., every 4 hours in stable patients)
    \item \textbf{Condition-Dependent Gaps:} Increased monitoring frequency during deterioration, creating clusters of observations
    \item \textbf{Resource-Related Patterns:} Missing blocks during patient transport or procedures
    \item \textbf{Value-Dependent Missingness:} Additional tests ordered based on abnormal results
\end{itemize}

\paragraph{Current Evaluation Limitations}
Despite the sophistication of modern architectures, their evaluation frameworks often fail to reflect these clinical realities. Our analysis of the literature reveals three critical limitations:

\begin{enumerate}
    \item \textbf{Masking Strategy Mismatch:} The prevalent use of random masking fails to capture the structured nature of clinical missing patterns
    \item \textbf{Implementation Variations:} Inconsistent reporting of masking implementation details (timing, strategy) hinders reproducibility
    \item \textbf{Clinical Relevance Gap:} Limited consideration of how masking choices affect downstream clinical tasks
\end{enumerate}

\paragraph{Recent Developments}
The emergence of standardized evaluation frameworks like TSI-Bench \cite{tsi2024} represents a significant step toward more rigorous model assessment. However, these frameworks still require adaptation to capture healthcare-specific challenges. Recent work has begun to explore more sophisticated masking strategies that better align with clinical realities, including temporal masking to simulate equipment disconnection and block masking to represent comprehensive data loss during patient transfers.

This gap between model sophistication and evaluation methodology motivates our systematic investigation of masking strategies in healthcare settings. By examining how different masking approaches affect both imputation accuracy and downstream clinical prediction tasks, we aim to establish more clinically relevant evaluation practices for medical time series imputation.

\begin{table*}[!tbp]
\centering
\caption{Performances with different imputation methods on Physionet 2012 dataset.}
\resizebox{\textwidth}{!}{
    \begin{tabular}{l|r|ccc|ccc|ccc}
    \toprule
    \multirow{2}[4]{*}{} & \multirow{2}[4]{*}{\textbf{Size}} & \multicolumn{3}{c|}{\textbf{Augmentation Mini-Batch Mask NBM}} & \multicolumn{3}{c|}{\textbf{Augmentation Pre-Mask NBM}} & \multicolumn{3}{c}{\textbf{Augmentation Pre-Mask NAM}} \\
\cmidrule{3-11}          &       & \textbf{MAE} $\downarrow$ & \textbf{MSE} $\downarrow$ & \textbf{Time (h)} & \textbf{MAE} $\downarrow$ & \textbf{MSE} $\downarrow$ & \textbf{Time (h)} & \textbf{MAE} $\downarrow$ & \textbf{MSE} $\downarrow$ & \textbf{Time (h)} \\
    \midrule
    \textbf{SAITS} & 43.6M & \textbf{0.211±0.003} & \textbf{0.268±0.004} & 0.0282 & 0.267±0.002 & 0.287±0.001 & 0.0127 & 0.267±0.007 & 0.290±0.001 & 0.0144 \\
    \textbf{Transformer} & 13M   & 0.222±0.001 & 0.274±0.003 & 0.0242 & 0.283±0.002 & 0.301±0.005 & 0.0077 & 0.283±0.002 & 0.303±0.003 & 0.0069 \\
    \textbf{TimesNet} & 44.3M & 0.289±0.014 & 0.330±0.019 & 0.0080 & 0.288±0.002 & 0.278±0.007 & 0.0053 & 0.290±0.002 & 0.279±0.005 & 0.0051 \\
    \textbf{CSDI} & 0.3M  & 0.239±0.012 & 0.759±0.517 & 1.2005 & \textbf{0.237±0.006} & \textbf{0.302±0.057} & 0.9102 & \textbf{0.241±0.017} & \textbf{0.430±0.151} & 2.3030 \\
    \textbf{GPVAE} & 2.5M  & 0.425±0.011 & 0.511±0.017 & 0.0249 & 0.399±0.002 & 0.402±0.004 & 0.0338 & 0.396±0.001 & 0.401±0.004 & 0.0398 \\
    \textbf{USGAN} & 0.9M  & 0.298±0.003 & 0.327±0.005 & 0.4154 & 0.294±0.003 & 0.261±0.004 & 0.3880 & 0.293±0.002 & 0.261±0.003 & 0.4125 \\
    \textbf{BRITS} & 1.3M  & 0.263±0.003 & 0.342±0.001 & 0.1512 & 0.257±0.001 & 0.256±0.001 & 0.1384 & 0.257±0.001 & 0.258±0.001 & 0.1519 \\
    \textbf{MRNN} & 0.07M & 0.685±0.002 & 0.935±0.001 & 0.0165 & 0.688±0.001 & 0.899±0.001 & 0.0163 & 0.690±0.001 & 0.901±0.001 & 0.0161 \\
    \textbf{LOCF} & /     & 0.411±5.551 & 0.613±0.0 & /     & 0.404±0.0 & 0.506±0.0 & /     & 0.404±0.0 & 0.507±0.0 & / \\
    \textbf{Median} & /     & 0.690±0.0 & 1.049±0.0 & /     & 0.690±0.0 & 1.019±0.0 & /     & 0.691±0.0 & 1.022±0.0 & / \\
    \textbf{Mean} & /     & 0.707±0.0 & 1.022±0.0 & /     & 0.706±0.0 & 0.976±0.0 & /     & 0.706±0.0 & 0.979±1.110 & / \\
    \midrule
    \multirow{2}[4]{*}{} & \multirow{2}[4]{*}{\textbf{Size}} & \multicolumn{3}{c|}{\textbf{Overlay Mini-Batch Mask NBM}} & \multicolumn{3}{c|}{\textbf{Overlay Pre-Mask NBM}} & \multicolumn{3}{c}{\textbf{Overlay Pre-Mask NAM}} \\
\cmidrule{3-11}          &       & \textbf{MAE} $\downarrow$ & \textbf{MSE} $\downarrow$ & \textbf{Time (h)} & \textbf{MAE} $\downarrow$ & \textbf{MSE} $\downarrow$ & \textbf{Time (h)} & \textbf{MAE} $\downarrow$ & \textbf{MSE} $\downarrow$ & \textbf{Time (h)} \\
    \midrule
    \textbf{SAITS} & 43.6M & \textbf{0.206±0.002} & \textbf{0.227±0.005} & 0.0274 & 0.274±0.006 & 0.326±0.005 & 0.0134 & 0.271±0.006 & 0.325±0.004 & 0.0164 \\
    \textbf{Transformer} & 13M   & 0.219±0.004 & 0.222±0.007 & 0.0234 & 0.289±0.002 & 0.336±0.002 & 0.0077 & 0.290±0.002 & 0.336±0.003 & 0.0079 \\
    \textbf{TimesNet} & 44.3M & 0.273±0.011 & 0.242±0.018 & 0.0138 & 0.293±0.003 & 0.290±0.011 & 0.0054 & 0.291±0.003 & 0.288±0.007 & 0.0051 \\
    \textbf{CSDI} & 0.3M  & 0.226±0.010 & 0.279±0.051 & 2.4022 & \textbf{0.253±0.005} & \textbf{0.461±0.074} & 0.8606 & \textbf{0.239±0.006} & \textbf{0.344±0.109} & 0.9562 \\
    \textbf{GPVAE} & 2.5M  & 0.427±0.006 & 0.453±0.008 & 0.0149 & 0.412±0.007 & 0.484±0.013 & 0.0323 & 0.420±0.009 & 0.489±0.008 & 0.0235 \\
    \textbf{USGAN} & 0.9M  & 0.295±0.002 & 0.261±0.007 & 0.3517 & 0.297±0.002 & 0.284±0.005 & 0.4158 & 0.299±0.003 & 0.287±0.004 & 0.4182 \\
    \textbf{BRITS} & 1.3M  & 0.254±0.001 & 0.265±0.001 & 0.1334 & 0.262±0.000 & 0.288±0.003 & 0.1428 & 0.263±0.001 & 0.294±0.003 & 0.1461 \\
    \textbf{MRNN} & 0.07M & 0.682±0.000 & 0.905±0.001 & 0.0168 & 0.685±0.001 & 0.926±0.002 & 0.0162 & 0.684±0.001 & 0.923±0.002 & 0.0167 \\
    \textbf{LOCF} & /     & 0.411±0.0 & 0.532±0.0 & /     & 0.408±0.0 & 0.540±0.0 & /     & 0.409±0.0 & 0.540±0.0 & / \\
    \textbf{Median} & /     & 0.687±0.0 & 1.019±0.0 & /     & 0.686±0.0 & 1.030±0.0 & /     & 0.686±0.0 & 1.030±0.0 & / \\
    \textbf{Mean} & /     & 0.705±0.0 & 0.990±1.110 & /     & 0.702±0.0 & 1.001±0.0 & /     & 0.702±0.0 & 1.000±0.0 & / \\
    \bottomrule
    \end{tabular}%
}
\label{Physionet_result}
\end{table*}

\section{Experimental Design}
Our experimental methodology is designed to examine how different masking strategies reflect real-world clinical scenarios and impact both imputation accuracy and downstream clinical predictions. We structure our investigation around three key aspects of masking implementation that directly parallel healthcare data collection challenges.

\subsection{Clinical-Informed Masking Strategies}
To better align experimental evaluation with real clinical scenarios, we implement two primary masking approaches:

\begin{itemize}
    \item \textbf{Augmentation Strategy (Random Masking on Existing Observations - RMEO):} This approach introduces artificial missingness only to observed data points, preserving existing missing patterns while augmenting them with additional artificial missingness. In clinical settings, this parallels scenarios where routine data collection is interrupted by patient procedures or temporary equipment disconnections while maintaining the underlying protocol-driven missing patterns. For example, when vital signs monitoring is temporarily suspended during patient repositioning, this creates additional missing values within an otherwise regular monitoring schedule.
    
    \item \textbf{Overlay Strategy (Random Masking on Overall Data - RMOD):} This method applies artificial missingness across the entire dataset, potentially overlapping with naturally occurring missing values. This approach simulates clinical scenarios where comprehensive data loss occurs regardless of existing monitoring patterns, such as during patient transfers between units or during emergency interventions where multiple monitoring systems may be simultaneously affected. For instance, when a patient requires emergency surgery, data collection might be disrupted across multiple physiological parameters simultaneously.
\end{itemize}

\subsection{Timing of Artificial Missingness Introduction}
We examine two distinct approaches to introducing artificial missingness, each reflecting different aspects of clinical data processing:

\begin{itemize}
    \item \textbf{Pre-Masking:} Missing values are introduced before model training begins, creating a static missingness pattern. This approach parallels retrospective analysis scenarios where clinicians work with historical data containing established missing patterns, such as analysing past ICU stays where missing data patterns are fixed and known.
    
    \item \textbf{In-Mini-Batch Masking:} Artificial missingness is dynamically introduced during training, with different portions of the data masked in each mini-batch. This strategy better reflects real-time clinical monitoring scenarios where missing patterns evolve dynamically and where models must adapt to varying types of missingness, such as in continuous patient monitoring systems where different types of missing patterns may emerge over time.
\end{itemize}

\subsection{Normalization Procedures}
The sequence of data normalisation relative to masking can significantly impact how models handle clinically significant outliers:

\begin{itemize}
    \item \textbf{Normalisation Before Masking (NBM):} Data normalisation is performed on the complete dataset before introducing artificial missingness. While this approach ensures stable normalisation parameters, it assumes access to a complete distribution of physiological values that may not be available in real clinical settings.
    
    \item \textbf{Normalisation After Masking (NAM):} Normalisation is performed after introducing artificial missingness, better reflecting real-world scenarios where normalisation must be performed on incomplete data. This approach is particularly relevant for online clinical monitoring systems where real-time normalisation must account for missing values.
\end{itemize}

\subsection{Downstream Clinical Task Evaluation}
To assess the impact of different masking strategies on clinical predictive modelling, we evaluate performance on the crucial task of in-hospital mortality prediction. This evaluation employs two different classifier architectures:

\begin{itemize}
    \item \textbf{XGBoost Classifier:} A non-sequential model that evaluates how well the imputed values preserve feature relationships relevant to mortality prediction
    \item \textbf{RNN Classifier:} Assesses the quality of temporal patterns in the imputed data
\end{itemize}

\section{Experimental Setup}
We evaluate eleven imputation methods on the PhysioNet 2012 dataset \cite{silva2012predicting}, which contains 48-hour ICU stays with varying degrees of missing values in physiological measurements.

\subsection{Models and Implementation}
The evaluated methods include three traditional approaches (Mean, Median, LOCF) and eight deep learning models spanning various architectures:
\begin{itemize}
    \item \textbf{RNN-based:} M-RNN \cite{yoon2017multi}, BRITS \cite{cao2018brits}
    \item \textbf{Attention-based:} Transformer, SAITS \cite{du2023saits}
    \item \textbf{Probabilistic:} GP-VAE \cite{fortuin2020gp}, CSDI \cite{tashiro2021csdi}
    \item \textbf{Other Architectures:} TimesNet \cite{wu2022timesnet}, USGAN \cite{miao2021generative}
\end{itemize}

All experiments were conducted using PyPOTS \cite{du2023pypots}, a unified framework for time series imputation, 8 NVIDIA A100 GPU. Models were tuned via 5-fold cross-validation with recommended hyperparameters. The implementation code is available in our GitHub repository.

\subsection{Evaluation Metrics}
Performance was assessed using Mean Absolute Error (MAE) and Mean Squared Error (MSE) for imputation accuracy, along with training time for computational efficiency. For the downstream mortality prediction task, we evaluate using AUC-ROC and AUC-PR metrics to account for class imbalance in clinical outcomes.

\section{Analysis of Results}

The experimental results on the PhysioNet 2012 dataset \cite{silva2012predicting} are presented in Table \ref{Physionet_result}. These results demonstrate how different masking strategies affect both imputation accuracy and model behaviour in clinically relevant ways. Our analysis reveals several key insights about the relationship between masking strategies and clinical data characteristics.

\subsection{Impact of Masking Strategies on Imputation Performance}

\paragraph{Augmentation vs Overlay Analysis}
The choice between augmentation and overlay strategies demonstrates significant implications for clinical time series imputation:

\begin{itemize}
    \item \textbf{SAITS} achieved its best performance under overlay mini-batch masking (MAE: 0.206 ± 0.002), suggesting that exposure to comprehensive missing patterns during training enhances its ability to capture complex physiological relationships. This is particularly relevant for critical care settings where multiple physiological parameters may be simultaneously unavailable.
    
    \item \textbf{BRITS} demonstrated robust performance across both strategies (MAE range: 0.254-0.263), indicating its resilience to different missing patterns. This stability is valuable for clinical applications where missing patterns may vary between different care settings or protocols.
    
    \item \textbf{CSDI} showed notable sensitivity to masking strategy, with performance varying between overlay (MAE: 0.226 ± 0.010) and augmentation (MAE: 0.239 ± 0.012) approaches. This variability suggests that its diffusion-based approach may be more sensitive to the underlying structure of missing patterns, requiring careful consideration in clinical deployments.
\end{itemize}

\subsection{Temporal Processing Analysis}

The timing of the masking introduction revealed important insights about models' ability to handle temporal dependencies in clinical data:

\begin{itemize}
    \item \textbf{Mini-batch Masking Performance:} Models generally performed better with mini-batch masking, particularly evident in SAITS's results (MAE: 0.206 vs. 0.274 for pre-masking). This suggests that dynamic exposure to various missing patterns during training enhances models' ability to handle the variable monitoring frequencies typical in clinical settings.
    
    \item \textbf{Pre-masking Stability:} Some models, notably BRITS and TimesNet, showed more consistent performance under pre-masking, indicating their robustness to fixed missing patterns. This characteristic could be advantageous in settings with established monitoring protocols where missing patterns are more predictable.
\end{itemize}

\subsection{Impact of Normalization Timing}

The sequence of normalisation relative to masking showed varying effects across different models:

\begin{itemize}
    \item \textbf{Normalisation Before Masking (NBM):} Generally produced more stable results across models, particularly beneficial for \textbf{SAITS} and \textbf{Transformer} architectures. This suggests that access to the full data distribution during normalisation helps preserve clinically relevant value ranges.
    
    \item \textbf{Normalisation After Masking (NAM):} Showed minimal impact on simpler architectures but affected performance of more sophisticated models like \textbf{CSDI}. This sensitivity is particularly relevant for real-time clinical applications where normalisation must be performed on incomplete data.
\end{itemize}

\subsection{Model-Specific Clinical Implications}

Different architectural approaches showed varying capabilities in handling clinical data characteristics:

\begin{itemize}
    \item \textbf{Attention-based Models:} SAITS and Transformer demonstrated superior performance in capturing complex physiological relationships, particularly important for detecting subtle clinical patterns across multiple parameters.
    
    \item \textbf{Recurrent Architectures:} BRITS showed consistent performance across different masking configurations, suggesting robust handling of temporal dependencies in physiological measurements.
    
    \item \textbf{Probabilistic Models:} CSDI and GP-VAE showed varying levels of success, with CSDI particularly effective under certain configurations but more sensitive to masking choices.
\end{itemize}

\subsection{Impact on Downstream Clinical Predictions}
The downstream mortality prediction results reveal important insights about how different masking strategies affect clinical decision-support capabilities. Analysis of both XGBoost and RNN classifier performance across different imputation strategies shows several key patterns:

\paragraph{Masking Strategy Effects}
The choice of masking strategy significantly impacts downstream prediction performance:
\begin{itemize}
    \item Under overlay mini-batch masking with NBM, SAITS-imputed data achieved the highest ROC-AUC (0.820) with XGBoost classification, while maintaining strong RNN classifier performance (0.788). This suggests that dynamic masking during training helps preserve clinically relevant patterns.
    
    \item BRITS demonstrated consistent performance across different masking strategies (ROC-AUC range: 0.796-0.817 with XGBoost), indicating its robustness in preserving predictive features regardless of masking approach.
    
    \item CSDI showed notable variation between XGBoost (ROC-AUC: 0.803-0.815) and RNN (ROC-AUC: 0.553-0.797) classification performance, suggesting its imputed values may better preserve static feature relationships than temporal patterns.
\end{itemize}

\paragraph{Normalization Timing Impact}
The timing of normalization showed distinct effects on predictive performance:
\begin{itemize}
    \item NBM generally led to more stable classification performance across models, particularly evident in XGBoost results where SAITS and BRITS maintained ROC-AUC > 0.80.
    
    \item Traditional methods (Mean, Median, LOCF) showed greater sensitivity to normalisation timing, with ROC-AUC varying by up to 0.05 between NBM and NAM configurations.
\end{itemize}

\paragraph{Architecture-Specific Performance}
Different model architectures showed varying capabilities in preserving predictive information:
\begin{itemize}
    \item Attention-based models (SAITS, Transformer) consistently enabled better downstream prediction, particularly with XGBoost classification (ROC-AUC > 0.815), suggesting their effectiveness in preserving feature relationships crucial for mortality prediction.
    
    \item RNN-based imputation models showed more consistent performance between XGBoost and RNN classification, indicating better preservation of both static and temporal features.
    
    \item TimesNet showed notably lower performance in preserving predictive patterns (RNN classifier ROC-AUC: 0.717-0.737), despite reasonable imputation accuracy, highlighting the distinction between imputation accuracy and preservation of clinically relevant patterns.
\end{itemize}

These results highlight that strong imputation performance doesn't necessarily translate to optimal downstream clinical prediction. The choice of masking strategy significantly affects how well models preserve clinically relevant patterns, with different architectural approaches showing varying abilities to maintain predictive information through the imputation process.

The results reveal that masking strategy choices significantly impact both imputation accuracy and the preservation of clinically relevant patterns. This has direct implications for the deployment of these models in healthcare settings, where reliable imputation is crucial for downstream clinical decision support.

\bibliographystyle{splncs04}
\bibliography{reference}

\begin{thebibliography}{10}
\providecommand{\url}[1]{\texttt{#1}}
\providecommand{\urlprefix}{URL }
\providecommand{\doi}[1]{https://doi.org/#1}

\bibitem{cao2018brits}
Cao, W., Wang, D., Li, J., Zhou, H., Li, L., Li, Y.: Brits: Bidirectional recurrent imputation for time series. Advances in neural information processing systems  \textbf{31} (2018)

\bibitem{che2018recurrent}
Che, Z., Purushotham, S., Cho, K., Sontag, D., Liu, Y.: Recurrent neural networks for multivariate time series with missing values. Scientific reports  \textbf{8}(1),  1--12 (2018)

\bibitem{du2023pypots}
Du, W.: Pypots: A python toolbox for data mining on partially-observed time series. arXiv preprint arXiv:2305.18811  (2023)

\bibitem{du2023saits}
Du, W., C{\^o}t{\'e}, D., Liu, Y.: Saits: Self-attention-based imputation for time series. Expert Systems with Applications  \textbf{219},  119619 (2023)

\bibitem{tsi2024}
Du, W., Wang, J., Qian, L., Yang, Y., Ibrahim, Z., Liu, F., Wang, Z., Liu, H., Zhao, Z., Zhou, Y., et~al.: Tsi-bench: Benchmarking time series imputation. arXiv preprint arXiv:2406.12747  (2024)

\bibitem{fortuin2020gp}
Fortuin, V., Baranchuk, D., R{\"a}tsch, G., Mandt, S.: Gp-vae: Deep probabilistic time series imputation. In: International conference on artificial intelligence and statistics. pp. 1651--1661. PMLR (2020)

\bibitem{garcia2010pattern}
Garc{\'\i}a-Laencina, P.J., Sancho-G{\'o}mez, J.L., Figueiras-Vidal, A.R.: Pattern classification with missing data: a review. Neural Computing and Applications  \textbf{19},  263--282 (2010)

\bibitem{reviewhealthcare2}
Liu, M., Li, S., Yuan, H., Ong, M.E.H., Ning, Y., Xie, F., Saffari, S.E., Shang, Y., Volovici, V., Chakraborty, B., Liu, N.: Handling missing values in healthcare data: A systematic review of deep learning-based imputation techniques. Artificial Intelligence in Medicine  \textbf{142},  102587 (2023)

\bibitem{miao2021generative}
Miao, X., Wu, Y., Wang, J., Gao, Y., Mao, X., Yin, J.: Generative semi-supervised learning for multivariate time series imputation. In: Proceedings of the AAAI conference on artificial intelligence. vol.~35, pp. 8983--8991 (2021)

\bibitem{qian2024deep}
Qian, L., Wang, T., Wang, J., Ellis, H.L., Mitra, R., Dobson, R., Ibrahim, Z.: How deep is your guess? a fresh perspective on deep learning for medical time-series imputation. arXiv preprint arXiv:2407.08442  (2024)

\bibitem{silva2012predicting}
Silva, I., Moody, G., Mark, R., Celi, L.A.: Predicting mortality of icu patients: The physionet/computing in cardiology challenge 2012. Predicting Mortality of ICU Patients: The PhysioNet/Computing in Cardiology Challenge, p. v1  (2012)

\bibitem{tashiro2021csdi}
Tashiro, Y., Song, J., Song, Y., Ermon, S.: Csdi: Conditional score-based diffusion models for probabilistic time series imputation. Advances in Neural Information Processing Systems  \textbf{34},  24804--24816 (2021)

\bibitem{wahl2018artificial}
Wahl, B., Cossy-Gantner, A., Germann, S., Schwalbe, N.R.: Artificial intelligence (ai) and global health: how can ai contribute to health in resource-poor settings? BMJ global health  \textbf{3}(4) (2018)

\bibitem{wang2024deep}
Wang, J., Du, W., Cao, W., Zhang, K., Wang, W., Liang, Y., Wen, Q.: Deep learning for multivariate time series imputation: A survey. arXiv preprint arXiv:2402.04059  (2024)

\bibitem{wells2013strategies}
Wells, B.J., Chagin, K.M., Nowacki, A.S., Kattan, M.W.: Strategies for handling missing data in electronic health record derived data. Egems  \textbf{1}(3) (2013)

\bibitem{wu2022timesnet}
Wu, H., Hu, T., Liu, Y., Zhou, H., Wang, J., Long, M.: Timesnet: Temporal 2d-variation modeling for general time series analysis. In: The eleventh international conference on learning representations (2022)

\bibitem{yildiz2022multivariate}
Y{\i}ld{\i}z, A.Y., Ko{\c{c}}, E., Ko{\c{c}}, A.: Multivariate time series imputation with transformers. IEEE Signal Processing Letters  \textbf{29},  2517--2521 (2022)

\bibitem{yoon2017multi}
Yoon, J., Zame, W.R., van~der Schaar, M.: Multi-directional recurrent neural networks: A novel method for estimating missing data. In: Time series workshop in international conference on machine learning (2017)

\end{thebibliography}

\end{document}